\documentclass[conference]{IEEEtran}
\ifCLASSINFOpdf
  \usepackage[pdftex]{graphicx}
  \graphicspath{{../pdf/}{../jpeg/}}
  \DeclareGraphicsExtensions{.pdf,.jpeg,.png}
\else
  \usepackage[dvips]{graphicx}
  \graphicspath{{../eps/}}
\fi
\usepackage{mathtools}
\usepackage{amsmath}

\DeclarePairedDelimiter\floor{\lfloor}{\rfloor}
\usepackage{color,soul}

\makeatletter
\def\ps@IEEEtitlepagestyle{
  \def\@oddfoot{\mycopyrightnotice}
  \def\@evenfoot{}
}
\def\mycopyrightnotice{
  {\footnotesize
  \begin{minipage}{\textwidth}
  \centering
  Copyright~\copyright~2020 IEEE. Personal use of this material is permitted. However, permission to use this  \\ 
  material for any other purposes must be obtained from the IEEE by sending a request to pubs-permissions@ieee.org. \\
  This paper has been accepted at ISCAS 2020 for publication.
  \end{minipage}
  }
}

\begin{document}

\newcommand{\archname}{\textbf{ESSOP} }
\newcommand{\archnamexy}{\textbf{ESSOP}}

\newcommand*{\red}{\textcolor{red}}
\newcommand*{\blue}{\textcolor{blue}}

%
\title{ESSOP: Efficient and Scalable Stochastic Outer Product Architecture for Deep Learning}



%
\author{\IEEEauthorblockN{Vinay Joshi\IEEEauthorrefmark{1}\IEEEauthorrefmark{2},
Geethan Karunaratne\IEEEauthorrefmark{1}\IEEEauthorrefmark{4},
Manuel Le Gallo\IEEEauthorrefmark{1}, 
Irem Boybat\IEEEauthorrefmark{1}\IEEEauthorrefmark{3},
Christophe Piveteau\IEEEauthorrefmark{1}\IEEEauthorrefmark{4},\\ 
Abu Sebastian\IEEEauthorrefmark{1},
Bipin Rajendran\IEEEauthorrefmark{5} and
Evangelos Eleftheriou\IEEEauthorrefmark{1} \\
Email: vinayjoshi.iitb@gmail.com, \{kar,anu,ibo,piv,ase,ele\}@zurich.ibm.com, bipin.rajendran@kcl.ac.uk}
\IEEEauthorblockA{\IEEEauthorrefmark{1}IBM Research - Zurich, 8803 R\"uschlikon, Switzerland}
\IEEEauthorblockA{\IEEEauthorrefmark{2}New Jersey Institute of Technology (NJIT), Newark, NJ 07102, USA}
\IEEEauthorblockA{\IEEEauthorrefmark{3}Ecole Polytechnique Federale de Lausanne (EPFL), 1015 Lausanne, Switzerland}
\IEEEauthorblockA{\IEEEauthorrefmark{4}ETH Z\"urich, 8092 Z\"urich, Switzerland}
\IEEEauthorblockA{\IEEEauthorrefmark{5}King's College London, Strand, London WC2R 2LS, United Kingdom}}


\maketitle

\begin{abstract}
Deep neural networks (DNNs) have surpassed human-level accuracy in a variety of cognitive tasks but at the cost of significant memory/time requirements in DNN training. This limits their deployment in  energy and memory limited applications that require real-time learning. Matrix-vector multiplications (MVM) and vector-vector outer product (VVOP) are the two most expensive operations associated with training of DNNs. Strategies to improve the efficiency of MVM computation in hardware have been demonstrated with minimal impact on training accuracy. However, the VVOP computation remains a relatively less explored bottleneck even with the aforementioned strategies. Stochastic computing (SC) has been proposed to improve the efficiency of VVOP computation but on relatively shallow networks with bounded activation functions and floating-point (FP) scaling of activation gradients. In this paper, we propose \archnamexy, an efficient and scalable stochastic outer product architecture based on the SC paradigm. We introduce efficient techniques to generalize SC for weight update computation in DNNs with the unbounded activation functions (e.g., ReLU), required by many state-of-the-art networks. Our architecture reduces the computational cost by re-using random numbers and replacing certain FP multiplication operations by bit shift scaling.
We show that the ResNet-32 network with 33 convolution layers and a fully-connected layer can be trained with \archname  on the CIFAR-10 dataset to achieve baseline comparable accuracy. Hardware design of \archname  at $14\,$nm technology node shows that, compared to a highly pipelined FP16 multiplier design, \archname  is $82.2\,$\% and $93.7\,$\% better in energy and area efficiency respectively for outer product computation.
\end{abstract}


%

\section{Introduction}
Research on developing accelerators for training deep neural networks (DNNs) has attracted significant interest. Several potential applications such as autonomous navigation, health care, and mobile devices require learning in-the-field while adhering to strict memory and energy budgets. DNN training demands significant time and compute/memory. The two most expensive computations in DNNs are the matrix-vector multiplications (MVM) and vector-vector outer product (VVOP) and both require $\mathcal{O}(N^{2})$ multiplications for a layer with a weight matrix of the size $N\times N$. 
Several strategies to improve efficiency of MVM computation have been proposed with minimal impact on the training accuracy.
These strategies leverage either low precision digital representation~\cite{binaryConnect,TWN} or crossbar architectures~\cite{Y2017burrAPX,ANT,vlsi_ase,Y2018sebastianJAP}. Less precise implementations of MVM are shown to perform sufficiently well for DNN training~\cite{binaryConnect,TWN,brein,mpa,BNNACC,rpu}.

For improving the efficiency of VVOP computation to calculte the weight updates, the algorithmic ideas that have been proposed so far require expensive multiplier circuits~\cite{binarizedNN,NNwFM,quantized1}.
Stochastic computing (SC) has been suggested as an efficient alternative to floating point (FP) multiplications, given that operands are real numbers in [$0, 1$].
 This poses challenges for DNN training as operations such as ReLu, batchnorm, etc. have unbounded outputs. Moreover, small error gradients are often quantized to 0 due to the limited precision in the range [0,1].
\setlength{\parskip}{0pt}

The main contributions of this paper are the following: (1) We propose an SC-based efficient architecture \archname for computing weight updates for DNN training. (2) We introduce efficient schemes to generalize SC-based multiplier to unbounded activation functions (e.g. ReLU) that are essential for DNN training~\cite{resnet,densenet,inception}. (3) We show that these improvements have minimal effect on training accuracy of a deep convolution neural network (CNN). (4) Post place and route results at $14\,$nm CMOS show that \archname design is $82.2\,$\% and $93.7\,$\% better in energy and area efficiency respectively, compared to a highly pipelined FP16 multiplier design for outer product computation.

\section{Background and Motivation} \label{sec:background}
\subsection{Neural Network Training} \label{subsec:nntraining}
\begin{figure}
\centering
\includegraphics[width=3in]{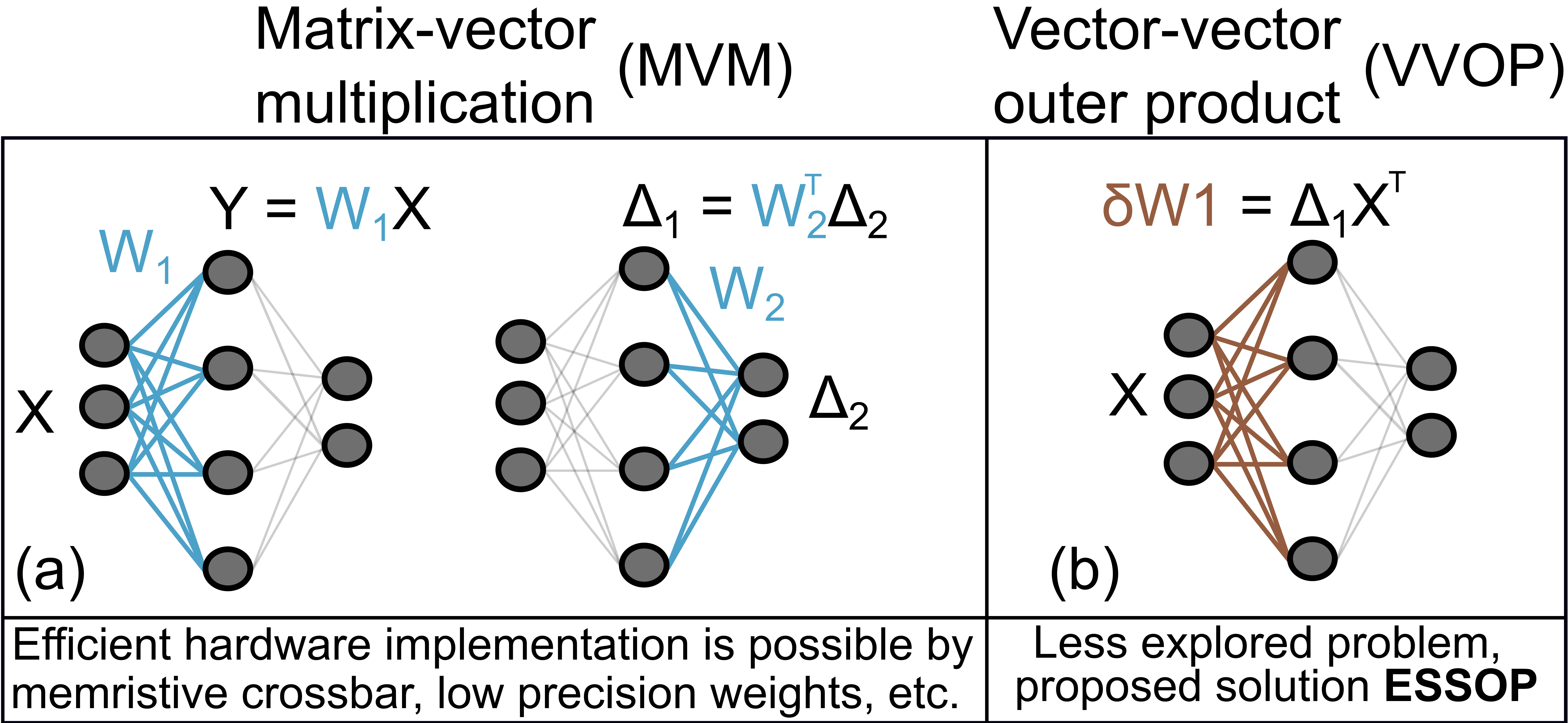}
\caption{Illustration of the two bottleneck operations in DNN training. Each of the illustrated operations require $\mathcal{O}(N^{2})$ multiplications. Unlike MVM, VVOP is a relatively less explored problem that eventually becomes the bottleneck even when MVM is efficiently implemented in hardware. The \archname architecture addresses this problem to enable efficient hardware implementation of  DNN training.} \label{fig:nn_training_ops}
\end{figure}

DNN training proceeds in three phases, namely (1) forward propagation, (2) backpropagation and (3) weight update. As shown in Fig.~\ref{fig:nn_training_ops},  MVM operation is essential in forward and backpropagation while during the  weight update phase, the  VVOP is computed between the error gradient $\Delta$ of that layer and the output activations of the previous layer $X$ to calculate the weight update matrix $\delta W$ (see Eq. (1)). Note that this $\delta W$ calculation, in general, applies to both fully-connected and convolution layers as well \cite{rpu_conv}. 
\begin{equation}
\delta W = \Delta  \times  X^{T}
\label{EQ:OUTERPRODUCT}
\end{equation}

\subsection{Stochastic Computing (SC)} \label{sec:stochastic_computing}
SC is a method of computing arithmetic operations using the  stochastic representation of real numbers constrained to the interval [$0, 1$], instead of using  real valued operands~\cite{sc3,sc2,sc1}. For notational convenience, we denote scalars by lowercase letters (e.g. $x$ $\in$ ${\rm I\!R}$) and vectors by uppercase letters
(e.g. $X$ $\in$ ${\rm I\!R^{N}}$).  To compute the stochastic representation $\Psi_{r}$ ($\in$ $\{0,1\}^{M}$) of a real number $r$, where $r\in[0, 1]$, a 
Bernoulli sequence with $M$ binary bits is computed such that the probability that any one of these bits being 1 is equal to r, i.e., $P(\Psi_{r}^{k}=1) = r  $ for  $k=1,2,\ldots, M$.

Using this representation, the product of two real numbers $c=r\times b$, where $b\in[0, 1]$, can be computed using the bit-wise AND
operation on the Bernoulli sequences as 
\begin{equation} \label{EQ:ber_and}
P(\Psi_{c}^{k} = 1) = P(\Psi_{r}^{k}=1) \wedge P(\Psi_{b}^{k}=1) \quad \forall k=1:M
\end{equation}
and
\begin{equation} \label{EQ:ber_and_avg}
c \sim E[\Psi_{c}] = E[\Psi_{r} \wedge \Psi_{b}]
\end{equation}
Equation (\ref{EQ:ber_and_avg}) thus replaces the expensive floating point multiplications with bitwise AND operations and subsequent summation
operations.  However, the range of numbers being multiplied in DNNs is usually not confined to [$0, 1$]. Equations (\ref{EQ:ber_and}) and (\ref{EQ:ber_and_avg}) can be generalized to numbers of arbitrary range; we illustrate this assuming $X$ and $\Delta$ both have $N$ elements and weight update is determined using VVOP computation as given by equation (\ref{EQ:OUTERPRODUCT}). Assuming that the vector $X$ lies in the range of $[-x_{max}, x_{max}]$ and similarly the  error gradient $\Delta$ lies in the range $[-\delta_{max}, \delta_{max}]$, we first normalize both $X$ and $\Delta$ vectors to constrain their values to [$-1, 1$] as,
\begin{align} \label{EQ:normalizationx}
\bar{X} &= \frac{X}{x_{max}} & \bar{\Delta} &= \frac{\Delta}{\delta_{max}}
\end{align}

Next, we denote the stochastic representation of all elements in a vector $X$ as $\Psi_{X}$ ($\in$ $\{0, 1\}^{N\times M}$). 
For computing the Bernoulli sequences of $\Psi_{\bar{X}}$ and $\Psi_{\bar{\Delta}}$ in hardware, we can implement a random number generator (RNG) to sample from the uniform distribution of [$0, 1$] and compare the normalized real number with the sampled random number:
\begin{equation} \label{EQ:berx}
\Psi_{\bar{\zeta}}^{i,k} = |\bar{\zeta}^{i}| \geq RNG_{\bar{\zeta}}^{k} \quad \forall k=1:M
\end{equation}
In  (\ref{EQ:berx}),  $\Psi_{\bar{\zeta}}^{i,k}$ is the $k^{th}$ Bernoulli event of the $i^{th}$ element of $\bar{\zeta}$ obtained by comparing $i^{th}$
element of $\bar{\zeta}$ with $k^{th}$ sample from corresponding random number generator $RNG_{\bar{\zeta}}^{k}$, where $\zeta$ is $X$ or $\Delta$.
We can approximate the product in equation (\ref{EQ:OUTERPRODUCT}) using SC as,
\begin{equation} \label{EQ:stochastic_mult}
\delta W^{ji}= sign(\delta W^{ji}) \times F_{scale}\times \sum_{k=1:M} \Psi_{\bar{\Delta}}^{j,k} \wedge \Psi_{\bar{X}}^{i,k}
\end{equation} where the parameter $F_{scale}$ is defined as $(x_{max} \times \delta_{max}/M)$.

From equations~(\ref{EQ:normalizationx})-(\ref{EQ:stochastic_mult}), it is clear that a SC-based multiplier implementation for VVOP
calculation presents the following challenges: (i) determination of the maximum elements $x_{max}$ and $\delta_{max}$ of the vectors $X$ and $\Delta$ respectively in  equation~(\ref{EQ:normalizationx}),  requiring $\mathcal{O}(N)$ floating point comparisons; (ii) floating point division for normalization in  equation~(\ref{EQ:normalizationx}) that requires $\mathcal{O}(N)$ floating point division operations; (iii) computation in  equation~(\ref{EQ:berx}) requires $\mathcal{O}(MN)$ random number generations; and (iv) scaling by $F_{scale}$ in  equation~(\ref{EQ:stochastic_mult}) requires $\mathcal{O}(N^{2})$ FP multiplication operations. We now discuss several techniques to address these challenges.  

\section{Optimization of the SC-based multiplier for the design of ESSOP architecture} \label{sec:optimization}
\subsection{Eliminating the normalization operations} \label{subsec:noNorm}
As discussed above, the normalization operation introduces $\mathcal{O}(N)$ floating point divisions. This can be addressed by the following improvement. Consider a number \textit{z} that lies in the range [$0, 1$] and another real number \textit{y} that is obtained by using
a constant positive scaling factor $y_{max}$ as $y = z \times y_{max}$. A Bernoulli representation ($\Psi_{z}^{k}$) of $z$ is obtained as $ z \geq RNG_{z}^{k}; \forall k=1:M$ and that of $y$ is,
\begin{equation} \label{EQ:Zber}
\Psi_{y}^{k} = [ y \geq y_{max} \times RNG_{z}^{k} ]  \quad \forall k=1:M
\end{equation}
In equation~\eqref{EQ:Zber}, RNG can be realized by using a linear feedback shift register (LFSR) circuit to generate $p$-bit pseudo-random numbers. Notably, the hardware realization of equation~(\ref{EQ:Zber}) does not require a multiplication with $y_{max}$ and can be realized by sampling few bits from the LFSR. For example, in a floating point representation, $2^{th}$ power in   $y_{max}$ can be used as an exponent and $RNG_{z}^{k}$ as mantissa to compute $y_{max} \times RNG_{z}^{k}$ without any floating point multiplications. Alternatively, in a fixed point representation, only a fraction of the bits generated from the LFSR need to be used to eliminate the fixed point multiplications in  equation~(\ref{EQ:Zber}). For instance, if $y_{max}$ requires only $8$-bits in the fixed point representation, $8$-bits could be sampled from the $p$-bit LFSR to compute $y_{max} \times RNG_{z}^{k}$. This  eliminates the need for $\mathcal{O}(N)$ expensive divisions irrespective of the numerical representation of $y$.

\subsection{Reusing the generated random numbers} \label{subsec:reuseRNG}
The next hurdle for a SC-based multiplier is the requirement to create $\mathcal{O}(MN)$ uniformly distributed $p$-bit random numbers. In order to efficiently utilize the generated random numbers, we propose to reuse the random numbers, for the computations in  equation~(\ref{EQ:berx}), $N$ times by generating only $M$ random numbers. 
Hence, we generate only $2M$ random numbers from the LFSR instead of $2NM$ random numbers for the $2N$ elements in $X$  and $\Delta$ combined. The first $M$ random numbers are used to generate the Bernoulli sequences of all the elements in the vector $X$ and the remaining $M$ random numbers are used to generate the Bernoulli sequences of all the elements in the vector $\Delta$. For example, to generate 8-bit long Bernoulli sequences corresponding to 256-element long vectors $X$  and $\Delta$, only $16$ random numbers are generated. The first 8 random numbers will be used to generate the Bernoulli sequence of all elements in the vector $X$  and the remaining $8$ for that of vector $\Delta$. 
With this modification, random number generation complexity is reduced to
$\mathcal{O}(M)$, making it independent of the dimensions of the weight matrix. 
As our detailed network simulations indicate, unintended correlations are not introduced by  reusing random numbers, as the two Bernoulli sequences in equation (\ref{EQ:stochastic_mult}) are uncorrelated.


\subsection{Approximating the scaling operations} \label{subsec:noScale}
Scaling the result of the AND operation in  equation~(\ref{EQ:stochastic_mult}) with $F_{scale}$ is an  $\mathcal{O}(N^2)$ operation involving
full-precision multiplications. To efficiently realize this scaling operation in  hardware, we propose to use the closest $2^{th}$ power of the number ${F}_{scale}$ obtained as shown below,
\begin{equation} \label{EQ:fscale2}
\widetilde{F}_{scale} =  2^{ \floor*{log_2({F}_{scale}) } } 
\end{equation}
where $ \floor*{x}$ denotes the largest integer smaller than $x$. Using $\widetilde{F}_{scale}$ instead of ${F}_{scale}$ makes the computations in
  equation~(\ref{EQ:stochastic_mult})  straightforward as only bit shift operations are required. Usually, $F_{scale}$ in DNNs will be smaller than 1 and
hence such a bit-shift operation will mostly be a right shift operation. In the case of stochastic gradient descent optimizer, note that the learning rate can also be accommodated in the $F_{scale}$ computation. 


\section{The ESSOP architecture}\label{sec:architecture}

\begin{figure}
\centering
\includegraphics[width=3in]{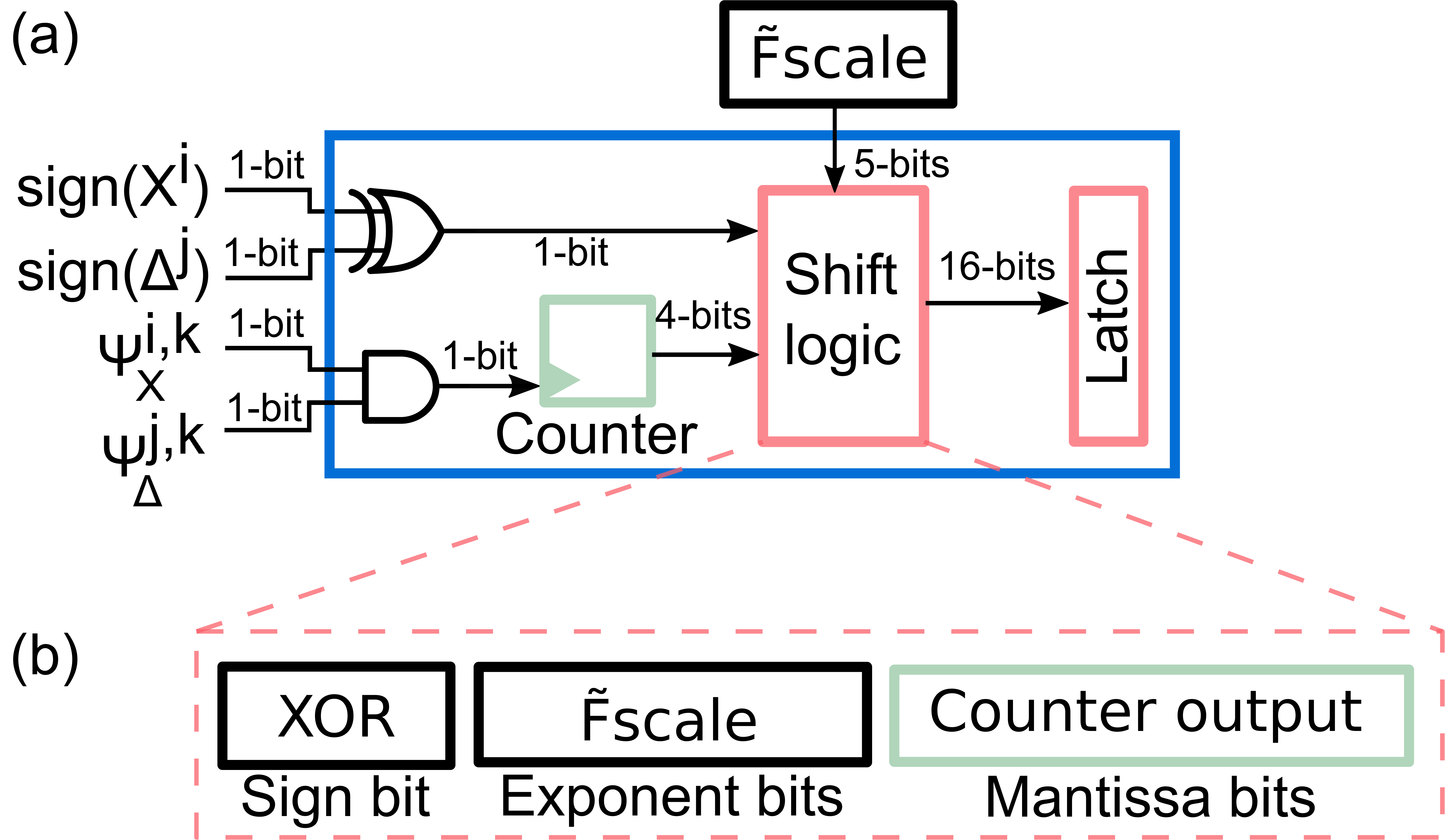}
\caption{The high-level design of a generalized \archname multiplier (unit cell) that can operate on two floating point numbers. (a) Internal blocks and the bit-lengths are shown for a unit cell implementation with FP16 inputs ($X$ and $\Delta$) and 16-bit sequence length. Superscirpts i, j denote the index of a real number in a vector and k denotes the index of a Bernoulli sample in a stochastic representation. (b) Example implementation of the shift logic for a floating point representation.} \label{fig:unitCell}
\end{figure}

\begin{figure}[b]
\centering
\includegraphics[width=2.5in]{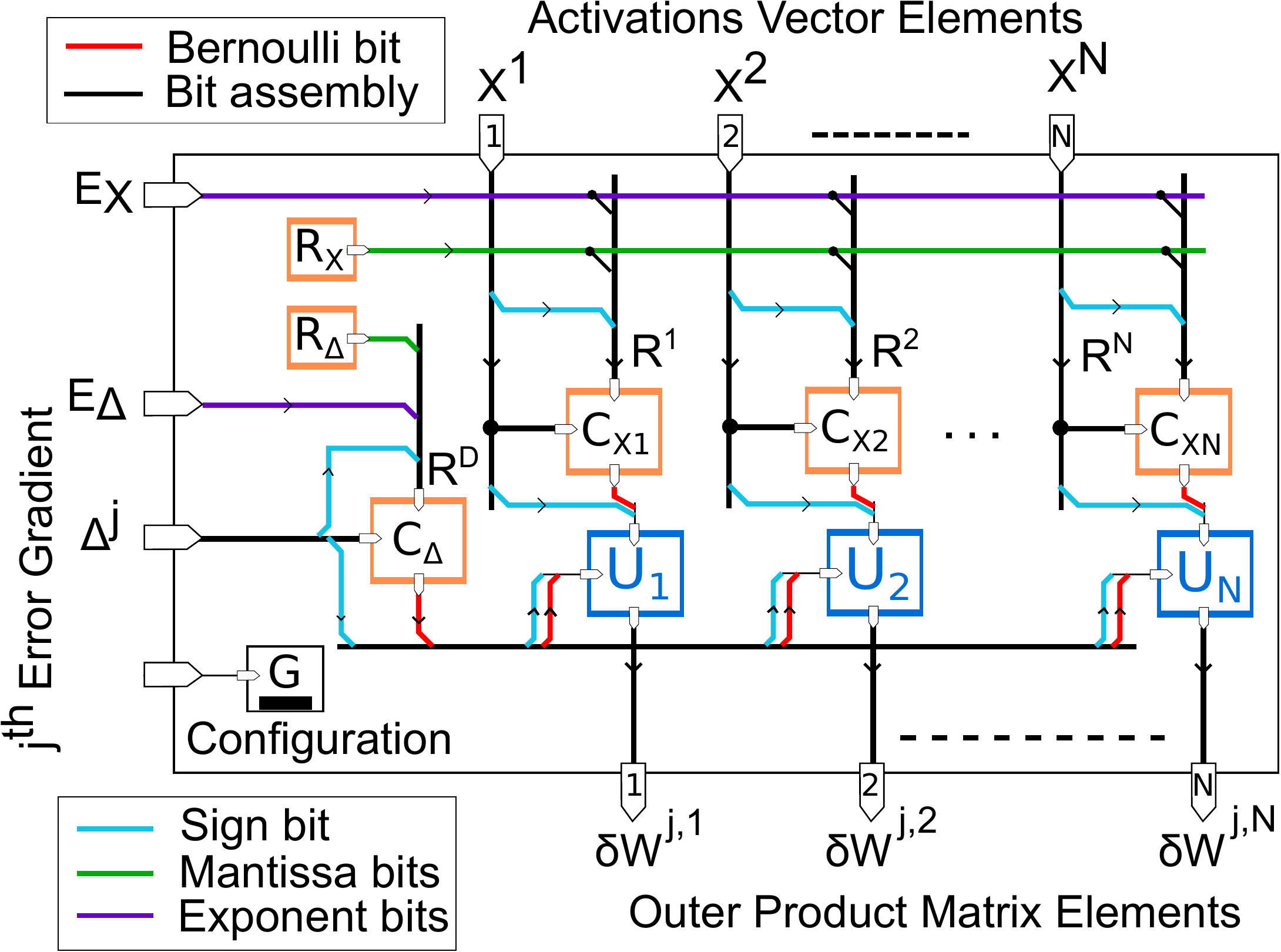}
\caption{The \archname architecture for computing the $N$ multiplications in parallel with the $N$ SC-based multipliers, assuming $X$ and $\Delta$ have floating point 16-bit precision. R, C and U stand for random number generator, comparator and unit cell respectively. This architecture can be reused to compute one full outer product. This also represent the high-level architecture used for our silicon design.} \label{fig:essop_arch}
\end{figure}

\subsection{Unit cell design}\label{subsec:unitcell}
We leverage innovations from Section \ref{sec:optimization} to develop the architecture of a single SC-based multiplier that we refer to as an \archname unit cell, shown in Fig.~\ref{fig:unitCell}(a). At the periphery of the unit cell, $M$-bit stochastic sequences $\psi^{i}_{X}$ and $\psi^{j}_{\Delta}$ are computed for two input real numbers $X^{i}$ and $\Delta^{j}$ respectively. The unit cell receives two inputs each with $2$ bits representation, with the first bit representing the stochastic representation of a real number and
the second bit is the sign of the real number. The sign of the final product is computed using a 1-bit XOR  on the sign bits of two real
numbers. 
In our design, we assume a simple $2$-input AND gate that is used $M$ times to compute the $M$-bit representation of   SC-based multiplication. For each cycle out of $M$ cycles, the output of the AND gate is fed to a counter that counts the number of $1$s in the resulting sequence. In the periphery of the unit cell, $\widetilde{F}_{scale}$ is computed, which is used by the shift logic circuit to scale the output of the counter to a desired range. Shift logic will depend on the digital representation used for the input real numbers. For example, for a floating point representation as shown in Fig.~\ref{fig:unitCell}(b), shift logic is as simple as copying the output of the XOR to the sign bit position, the counter output to the mantissa position and $\widetilde{F}_{scale}$ value to the exponent position. 
Finally, the result of the shift logic circuit is stored in a latch (or in a desired memory location) for the weight update. Note that  the SC-based multiplier requires only $M$ clock cycles to compute one product.

\subsection{The multi-cell architecture of ESSOP} \label{subsec:ESSOPengine}

To compute all the elements of an outer product matrix, either a unit cell can be multiplexed or multiple unit cells can be used simultaneously. The proposed \archname architecture has multiple unit cells stacked in a single row.  
The high-level architecture of \archname is shown in Fig.~\ref{fig:essop_arch}, and has $N$ unit cells (U) arranged in a row. 
At the periphery of the unit cells, all inputs have their corresponding comparator (C). Each comparator receives two inputs, one from either an element of an activation vector ($X$) or error gradient vector ($\Delta$), second from a corresponding random number generator (R). The comparator generates one stochastic bit by comparing two inputs at a time. In the $M$-bit implementation of the ESSOP, several such random numbers could be fed to the comparator circuit to compute $M$ comparisons in total, resulting in an $M$-bit long stochastic sequence.
As discussed in Section \ref{subsec:reuseRNG}, there are exactly two RNGs, one for $X$ and another for $\Delta$, that generate $M$ random numbers each for every outer product.

\section{Numerical validation} \label{sec:numerical_validation}

\begin{figure}
\centering
\includegraphics[width=3in]{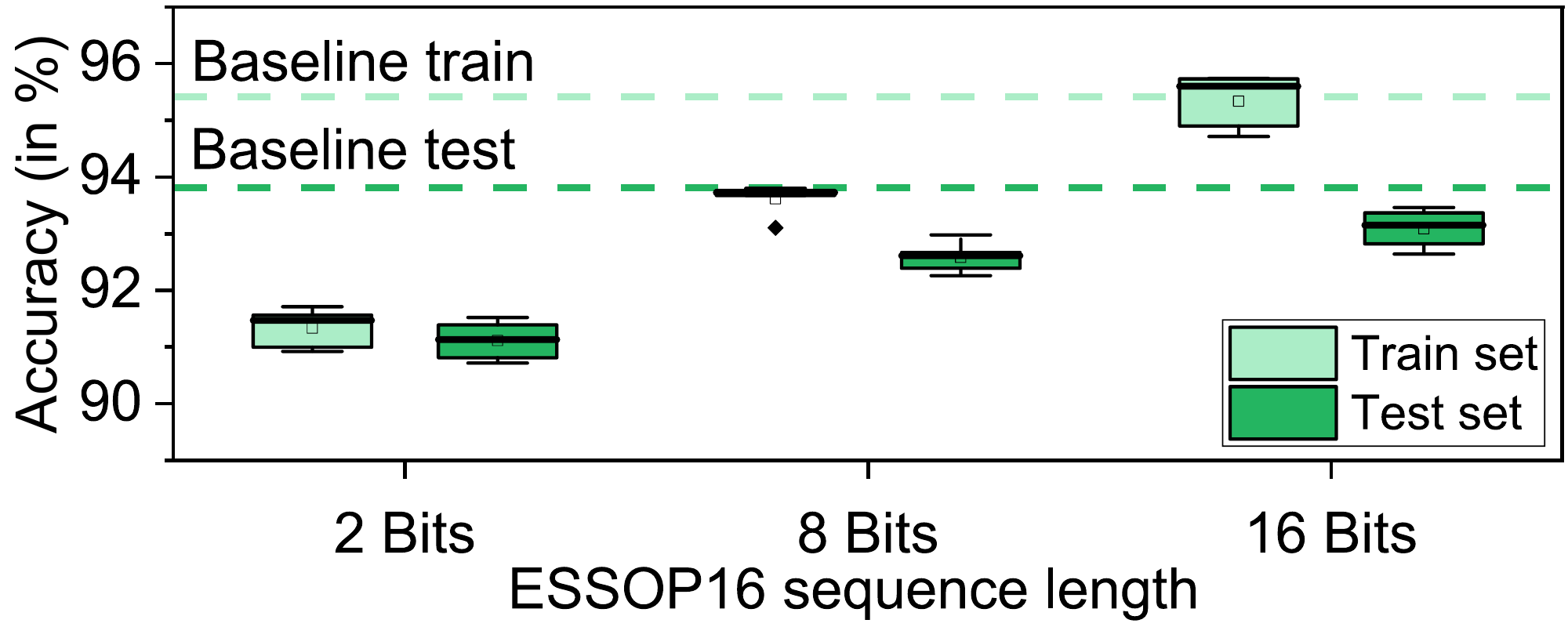}
\caption{The accuracy of the ResNet-32 network as a function \archname  with the  VVOP inputs represented using   16, 8, and 2-bits  long Bernoulli sequences.   16-bit sequence  is enough to achieve baseline comparable accuracy and  2-bit representation suffers only 2.6\% drop on an average compared to the baseline. Each box plot shows five independent runs with different seeds.}  \label{fig:resnet_training_curves}
\end{figure}


We train the ResNet-32\cite{resnet} network on CIFAR-10~\cite{cifar} dataset to validate the \archname architecture. We use \archname to compute weight updates in all the 33 convolution layers and in the final layer of the ResNet-32 network. The CIFAR-10 dataset has $50$K images in the training set and $10$K images in the test set. Each image in the dataset has a resolution of  $32\times32$ pixels and three channels and belongs one of the ten classes. We preprocess CIFAR-10 images by implementing the commonly used image processing steps for the family of residual networks as reported in \cite{cutout}. 
The simulation was performed with FP16 precision for data and computation, with a  mini-batch size of $100$ images  for $200$ epochs. We used initial learning rate (LR) of $0.1$ with LR evolution (LRE) for baseline as  in \cite{resnet}; LRE is tuned for better accuracy in \archname implementation. The categorical cross-entropy loss function is minimized using stochastic gradient descent with momentum of $0.9$. In our results, we denote \archnamexy$16$(M) to indicate FP16 precision for input operands ($X$ and $\Delta$) represented using $M$-bit Bernoulli sequences.
Fig.~\ref{fig:resnet_training_curves} shows the accuracy of ResNet-32 as a function of \archnamexy16 sequence length. The test accuracy drop with \archnamexy16(16) is only 0.25\% compared to the baseline. Experiments on different sequence lengths indicate that $16$-bits is sufficient to achieve  close to  baseline accuracy with FP16 outer product. \archnamexy16(16) shows on an average of 0.73\% drop in the test accuracy compared to the baseline accuracy.  \archnamexy16(8) has on an average of $1.13$\% drop in  test accuracy compared to the baseline. Remarkably, \archnamexy16(2) has on an average of only $2.6$\% drop in the test accuracy compared to the baseline. It is important to note that with a sequence length of $2$-bits, it is possible to compute the weight update in just $2$ clock cycles.

\section{Post layout results}

\begin{figure}
\centering
\includegraphics[width=2.5in]{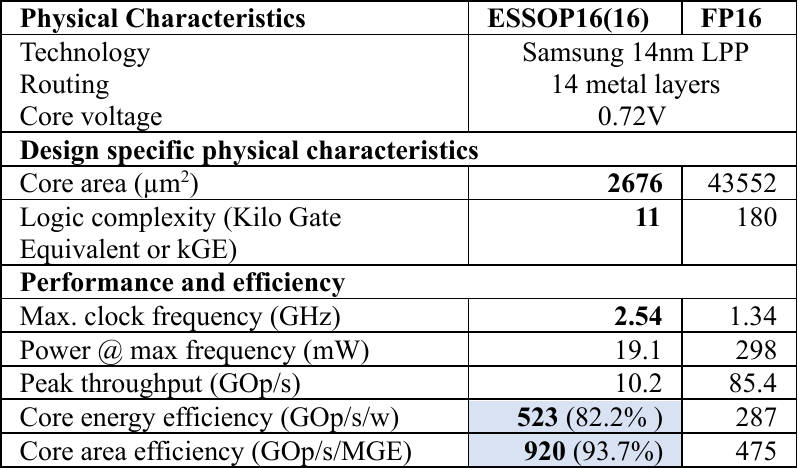}
\caption{Physical and performance characteristics of \archnamexy16(16) architecture vs floating point 16-bits (FP16) multiplier array. }
\label{tab:perform_spec}
\end{figure}

 In this section, we present the details of hardware implementation of \archnamexy16 and compare its post route layout performance with that of a highly pipelined FP16 multiplier design after place and route. FP16 design is an array of $N=64$  FP16 multipliers (from Samsung's Low Power Plus (LPP) library) that compute $64$ FP16 multiplications in parallel.   Similarly, \archnamexy16 design has   $64$ unit cells ($U_{1}$ to $U_{64}$) to compute $64$ elements of an outer product matrix in parallel, as illustrated in Fig.~\ref{fig:essop_arch}. Inputs $X$ and $\Delta$, which  are in FP16 precision,  are fed to the corresponding comparator circuit ($C_{1}$ to $C_{64}$). Two RNGs $R_{X}$ and $R_{\Delta}$ corresponding to $X$ and $\Delta$ generate $M$ random numbers each. Each RNG generates only mantissa part of the random number, exponent ($E_{X}$ or $E_{\Delta}$) is derived from the absolute maximum number in the vector $X$ or $\Delta$ and sign bit is derived from another input of a corresponding comparator. Configuration parameters such as Bernoulli sequence length is stored in configuration register G. Post place and route results at $14\,$nm using Samsung LPP libraries   are shown in Fig.~\ref{tab:perform_spec}. These results indicate \archnamexy16(16) design, even though sequential and not pipelined, operates at $1.9\times$ higher frequency and achieves $82.2\,$\% and $93.7\,$\% better energy and area efficiency respectively, compared to the FP16 multiplier array for outer product computation.

\section{Conclusion}
We proposed an efficient hardware architecture \archname that facilitates training of   deep neural networks. The central idea is to efficiently implement the vector-vector outer product calculation associated with the weight updates using stochastic computing. We proposed   efficient schemes to implement the stochastic computing-based multipliers that can generalize to operands with unbounded magnitude and significantly reduce the computational cost by re-using random numbers. This addresses a significant performance bottleneck for training DNNs in hardware, particularly for applications with stringent constraints on area and energy.   \archname complements architectures that accelerate matrix-vector multiply operations associated with the forward and backpropagations   where weights are represented in  low precision or stored in   computational memory based crossbar array architectures. We evaluated \archname on a 32-layer deep CNN that achieves baseline comaprable accuracy for a sequence length of $16$-bits.    $14\,$nm place and route of the \archname architecture compared with FP16 design shows $82.2\,$\% and $93.7\,$\% improvement in energy and area efficiency respectively for outer product computation.

\section*{Acknowledgment}
This project was supported partially by the Semiconductor Research Corporation.

\bibliographystyle{IEEEtran}
\bibliography{ref}

\begin{thebibliography}{10}
\providecommand{\url}[1]{#1}
\csname url@samestyle\endcsname
\providecommand{\newblock}{\relax}
\providecommand{\bibinfo}[2]{#2}
\providecommand{\BIBentrySTDinterwordspacing}{\spaceskip=0pt\relax}
\providecommand{\BIBentryALTinterwordstretchfactor}{4}
\providecommand{\BIBentryALTinterwordspacing}{\spaceskip=\fontdimen2\font plus
\BIBentryALTinterwordstretchfactor\fontdimen3\font minus
  \fontdimen4\font\relax}
\providecommand{\BIBforeignlanguage}[2]{{%
\expandafter\ifx\csname l@#1\endcsname\relax
\typeout{** WARNING: IEEEtran.bst: No hyphenation pattern has been}%
\typeout{** loaded for the language `#1'. Using the pattern for}%
\typeout{** the default language instead.}%
\else
\language=\csname l@#1\endcsname
\fi
#2}}
\providecommand{\BIBdecl}{\relax}
\BIBdecl

\bibitem{binaryConnect}
\BIBentryALTinterwordspacing
M.~Courbariaux, Y.~Bengio, and J.~David, ``Binaryconnect: Training deep neural
  networks with binary weights during propagations,'' \emph{CoRR}, vol.
  abs/1511.00363, 2015. [Online]. Available:
  \url{http://arxiv.org/abs/1511.00363}
\BIBentrySTDinterwordspacing

\bibitem{TWN}
\BIBentryALTinterwordspacing
F.~Li and B.~Liu, ``Ternary weight networks,'' \emph{CoRR}, vol.
  abs/1605.04711, 2016. [Online]. Available:
  \url{http://arxiv.org/abs/1605.04711}
\BIBentrySTDinterwordspacing

\bibitem{Y2017burrAPX}
G.~W. Burr, R.~M. Shelby, A.~Sebastian, S.~Kim, S.~Kim, S.~Sidler, K.~Virwani,
  M.~Ishii, P.~Narayanan, A.~Fumarola \emph{et~al.}, ``Neuromorphic computing
  using non-volatile memory,'' \emph{Advances in Physics: X}, vol.~2, no.~1,
  pp. 89--124, 2017.

\bibitem{ANT}
\BIBentryALTinterwordspacing
V.~Joshi, M.~L. Gallo, I.~Boybat, S.~Haefeli, C.~Piveteau, M.~Dazzi,
  B.~Rajendran, A.~Sebastian, and E.~Eleftheriou, ``Accurate deep neural
  network inference using computational phase-change memory,'' \emph{CoRR},
  vol. abs/1906.03138, 2019. [Online]. Available:
  \url{http://arxiv.org/abs/1906.03138}
\BIBentrySTDinterwordspacing

\bibitem{vlsi_ase}
A.~{Sebastian}, I.~{Boybat}, M.~{Dazzi}, I.~{Giannopoulos}, V.~{Jonnalagadda},
  V.~{Joshi}, G.~{Karunaratne}, B.~{Kersting}, R.~{Khaddam-Aljameh}, S.~R.
  {Nandakumar}, A.~{Petropoulos}, C.~{Piveteau}, T.~{Antonakopoulos},
  B.~{Rajendran}, M.~L. {Gallo}, and E.~{Eleftheriou}, ``Computational
  memory-based inference and training of deep neural networks,'' in \emph{2019
  Symposium on VLSI Technology}, June 2019, pp. T168--T169.

\bibitem{Y2018sebastianJAP}
A.~Sebastian, M.~Le~Gallo, G.~W. Burr, S.~Kim, M.~BrightSky, and
  E.~Eleftheriou, ``Tutorial: Brain-inspired computing using phase-change
  memory devices,'' \emph{Journal of Applied Physics}, vol. 124, no.~11, p.
  111101, 2018.

\bibitem{brein}
K.~{Ando}, K.~{Ueyoshi}, K.~{Orimo}, H.~{Yonekawa}, S.~{Sato}, H.~{Nakahara},
  S.~{Takamaeda-Yamazaki}, M.~{Ikebe}, T.~{Asai}, T.~{Kuroda}, and
  M.~{Motomura}, ``B{R}ein memory: A single-chip binary/ternary reconfigurable
  in-memory deep neural network accelerator achieving 1.4 {TOPS} at 0.6 {W},''
  \emph{IEEE Journal of Solid-State Circuits}, vol.~53, no.~4, pp. 983--994,
  April 2018.

\bibitem{mpa}
\BIBentryALTinterwordspacing
S.~R. Nandakumar, M.~L. Gallo, I.~Boybat, B.~Rajendran, A.~Sebastian, and
  E.~Eleftheriou, ``Mixed-precision training of deep neural networks using
  computational memory,'' \emph{CoRR}, vol. abs/1712.01192, 2017. [Online].
  Available: \url{http://arxiv.org/abs/1712.01192}
\BIBentrySTDinterwordspacing

\bibitem{BNNACC}
E.~{Nurvitadhi}, D.~{Sheffield}, {Jaewoong Sim}, A.~{Mishra}, G.~{Venkatesh},
  and D.~{Marr}, ``Accelerating binarized neural networks: Comparison of
  {FPGA}, {CPU}, {GPU}, and {ASIC},'' in \emph{2016 International Conference on
  Field-Programmable Technology (FPT)}, Dec 2016, pp. 77--84.

\bibitem{rpu}
\BIBentryALTinterwordspacing
G.~Tayfun and Y.~Vlasov, ``Acceleration of deep neural network training with
  resistive cross-point devices,'' \emph{CoRR}, vol. abs/1603.07341, 2016.
  [Online]. Available: \url{http://arxiv.org/abs/1603.07341}
\BIBentrySTDinterwordspacing

\bibitem{binarizedNN}
\BIBentryALTinterwordspacing
M.~Courbariaux and Y.~Bengio, ``Binarynet: Training deep neural networks with
  weights and activations constrained to +1 or -1,'' \emph{CoRR}, vol.
  abs/1602.02830, 2016. [Online]. Available:
  \url{http://arxiv.org/abs/1602.02830}
\BIBentrySTDinterwordspacing

\bibitem{NNwFM}
\BIBentryALTinterwordspacing
Z.~Lin, M.~Courbariaux, R.~Memisevic, and Y.~Bengio, ``Neural networks with few
  multiplications,'' \emph{CoRR}, vol. abs/1510.03009, 2015. [Online].
  Available: \url{http://arxiv.org/abs/1510.03009}
\BIBentrySTDinterwordspacing

\bibitem{quantized1}
V.~Vanhoucke, A.~Senior, and M.~Z. Mao, ``Improving the speed of neural
  networks on {CPU}s,'' in \emph{Deep Learning and Unsupervised Feature
  Learning Workshop, NIPS 2011}, 2011.

\bibitem{resnet}
\BIBentryALTinterwordspacing
K.~He, X.~Zhang, S.~Ren, and J.~Sun, ``Deep residual learning for image
  recognition,'' \emph{CoRR}, vol. abs/1512.03385, 2015. [Online]. Available:
  \url{http://arxiv.org/abs/1512.03385}
\BIBentrySTDinterwordspacing

\bibitem{densenet}
\BIBentryALTinterwordspacing
G.~Huang, Z.~Liu, and K.~Q. Weinberger, ``Densely connected convolutional
  networks,'' \emph{CoRR}, vol. abs/1608.06993, 2016. [Online]. Available:
  \url{http://arxiv.org/abs/1608.06993}
\BIBentrySTDinterwordspacing

\bibitem{inception}
\BIBentryALTinterwordspacing
C.~Szegedy, W.~Liu, Y.~Jia, P.~Sermanet, S.~E. Reed, D.~Anguelov, D.~Erhan,
  V.~Vanhoucke, and A.~Rabinovich, ``Going deeper with convolutions,''
  \emph{CoRR}, vol. abs/1409.4842, 2014. [Online]. Available:
  \url{http://arxiv.org/abs/1409.4842}
\BIBentrySTDinterwordspacing

\bibitem{rpu_conv}
\BIBentryALTinterwordspacing
T.~Gokmen, M.~Onen, and W.~Haensch, ``Training deep convolutional neural
  networks with resistive cross-point devices,'' \emph{Frontiers in
  Neuroscience}, vol.~11, p. 538, 2017. [Online]. Available:
  \url{https://www.frontiersin.org/article/10.3389/fnins.2017.00538}
\BIBentrySTDinterwordspacing

\bibitem{sc3}
\BIBentryALTinterwordspacing
A.~Alaghi and J.~P. Hayes, ``Survey of stochastic computing,'' \emph{ACM Trans.
  Embed. Comput. Syst.}, vol.~12, no.~2s, pp. 92:1--92:19, May 2013. [Online].
  Available: \url{http://doi.acm.org/10.1145/2465787.2465794}
\BIBentrySTDinterwordspacing

\bibitem{sc2}
\BIBentryALTinterwordspacing
B.~R. Gaines, ``Stochastic computing,'' in \emph{Proceedings of the April
  18-20, 1967, Spring Joint Computer Conference}, ser. AFIPS '67
  (Spring).\hskip 1em plus 0.5em minus 0.4em\relax New York, NY, USA: ACM,
  1967, pp. 149--156. [Online]. Available:
  \url{http://doi.acm.org/10.1145/1465482.1465505}
\BIBentrySTDinterwordspacing

\bibitem{sc1}
\BIBentryALTinterwordspacing
W.~J. Poppelbaum, C.~Afuso, and J.~W. Esch, ``Stochastic computing elements and
  systems,'' in \emph{Proceedings of the November 14-16, 1967, Fall Joint
  Computer Conference}, ser. AFIPS '67 (Fall).\hskip 1em plus 0.5em minus
  0.4em\relax New York, NY, USA: ACM, 1967, pp. 635--644. [Online]. Available:
  \url{http://doi.acm.org/10.1145/1465611.1465696}
\BIBentrySTDinterwordspacing

\bibitem{cifar}
A.~Krizhevsky, ``Learning multiple layers of features from tiny images,''
  \emph{University of Toronto}, 05 2012.

\bibitem{cutout}
\BIBentryALTinterwordspacing
T.~Devries and G.~W. Taylor, ``Improved regularization of convolutional neural
  networks with cutout,'' \emph{CoRR}, vol. abs/1708.04552, 2017. [Online].
  Available: \url{http://arxiv.org/abs/1708.04552}
\BIBentrySTDinterwordspacing

\end{thebibliography}
\end{document}